%% file: Main.tex
\documentclass[9pt,conference]{IEEEtran}

\usepackage{cite}
\usepackage{amsmath,amssymb,amsfonts}
\usepackage{algorithmic}
\usepackage{graphicx}
\usepackage{textcomp}
\usepackage[table]{xcolor}
\usepackage{multicol}
\usepackage{multirow}
\usepackage{makecell}
\usepackage{tablefootnote}
\usepackage{pifont}
\usepackage{courier} 
\usepackage{tikz}
\usepackage[hidelinks]{hyperref}
\usepackage{url}
\usepackage{listings}
\usepackage{courier} 
\usepackage{bm}
\usepackage{subcaption}

\definecolor{darkgreen}{rgb}{0.0, 0.6, 0.0}
\definecolor{darkskyblue}{rgb}{0.0, 0.45, 0.8}
\definecolor{lightpurple}{rgb}{0.6, 0.35, 0.9}

\def\BibTeX{{\rm B\kern-.05em{\sc i\kern-.025em b}\kern-.08em
    T\kern-.1667em\lower.7ex\hbox{E}\kern-.125emX}}
\begin{document}

\newcommand{\oursys}{ReChisel}

\newcommand{\ttf}[1]{\scriptsize\ttfamily #1}
\newcommand{\tcode}[1]{\textbf{\textcolor{darkgreen}{#1}}}
\newcommand{\wcode}[1]{\textbf{\textcolor{red}{#1}}}
\newcommand{\fdbk}[1]{\textbf{\textcolor{black}{#1}}}
\newcommand{\corr}[1]{\textbf{\textcolor{black}{#1}}}

\newcommand{\cmark}{\ding{51}} 
\newcommand{\xmark}{\ding{55}} 

\newcommand{\stepa}{\textbf{Step}~\ding{182}}
\newcommand{\stepb}{\textbf{Step}~\ding{183}}
\newcommand{\stepc}{\textbf{Step}~\ding{184}}
\newcommand{\stepd}{\textbf{Step}~\ding{185}}
\newcommand{\stepe}{\textbf{Step}~\ding{186}}
\newcommand{\stepf}{\textbf{Step}~\ding{187}}
\newcommand{\stepg}{\textbf{Step}~\ding{188}}

\title{\oursys: Effective Automatic Chisel Code Generation by LLM with Reflection
}

\DeclareRobustCommand{\IEEEauthorrefmark}[1]{\smash{\textsuperscript{\footnotesize #1}}}

\author{
\IEEEauthorblockN{
    Juxin Niu\IEEEauthorrefmark{1}, 
    Xiangfeng Liu\IEEEauthorrefmark{1,2}, 
    Dan Niu\IEEEauthorrefmark{3}, 
    Xi Wang\IEEEauthorrefmark{4,5}\IEEEauthorrefmark{*}, 
    Zhe Jiang\IEEEauthorrefmark{4,5}, 
    Nan Guan\IEEEauthorrefmark{1}\IEEEauthorrefmark{*}
}
\IEEEauthorblockA{\IEEEauthorrefmark{1} Department of Computer Science, City University of Hong Kong, Hong Kong SAR
}
\IEEEauthorblockA{\IEEEauthorrefmark{2} School of Computer Science and Engineering, Northeastern University, China
}
\IEEEauthorblockA{\IEEEauthorrefmark{3} School of Automation, Southeast University, China
}
\IEEEauthorblockA{\IEEEauthorrefmark{4} National Center of Technology Innovation for EDA, China \IEEEauthorrefmark{5} School of Integrated Circuits, Southeast University, China
}
\IEEEauthorblockA{
Email: 
juxin.niu@my.cityu.edu.hk, xiangfengliu.cn@gmail.com, 
101011786@seu.edu.cn, xi.wang@seu.edu.cn, \\ zhejiang.arch@gmail.com,
nanguan@cityu.edu.hk
}
\IEEEauthorblockA{\IEEEauthorrefmark{*}Corresponding authors.}
}


\maketitle

\begin{abstract}
Coding with hardware description languages (HDLs) such as Verilog is a time-intensive and laborious task. With the rapid advancement of large language models (LLMs), there is increasing interest in applying LLMs to assist with HDL coding. Recent efforts have demonstrated the potential of LLMs in translating natural language to traditional HDL Verilog.
Chisel, a next-generation HDL based on Scala, introduces higher-level abstractions, facilitating more concise, maintainable, and scalable hardware designs. 
However, the potential of using LLMs for Chisel code generation
remains largely unexplored.
This work proposes \oursys, an LLM-based agentic system designed to enhance the effectiveness of Chisel code generation.
\oursys\ incorporates a reflection mechanism to iteratively refine
the quality of generated code using feedback from compilation and
simulation processes, and introduces an escape mechanism to break
free from non-progress loops. 
Experiments demonstrate that \oursys\ significantly
improves the success rate of Chisel code generation, achieving performance comparable to state-of-the-art LLM-based agentic systems for Verilog code generation.

\end{abstract}


\input{z-introduction}

\input{z-background}
\input{z-motivation}

\input{z-new-design}

\input{z-evaluation}

\input{z-case-study}

\section{Conclusion}

In this work, we introduce \oursys, an LLM-based agentic system designed to enhance the effectiveness of Chisel code generation. 
\oursys\ incorporates a reflection mechanism to iteratively refine the quality of generated code using feedback from compilation and simulation processes, and introduces an escape mechanism to break free from non-progress loops.
Experiments across three benchmarks and five mainstream LLMs demonstrate that \oursys\ significantly improves the success rate of Chisel code generation, achieving performance comparable to state-of-the-art LLM-based agentic systems for Verilog code generation. 
This work highlights the potential of Chisel in LLM-assisted hardware design.

\section{Acknowledgement}

This work is supported by the National Key Research and Development Program (Grant No. 2024YFB4405600), the National Natural Science Foundation of China (Grant No. 62472086), the Basic Research Program of Jiangsu (Grants No. BK20243042, BG2024010), and the Start-up Research Fund of Southeast University (Grant No. RF1028624005).

\bibliographystyle{IEEEtranS}
\bibliography{ref}

\end{document}

%% file: z-introduction.tex
\section{Introduction}


In hardware design, coding with hardware description languages (HDLs) like Verilog is often a time-consuming and labor-intensive task. Large language models (LLMs) have demonstrated promising capabilities in assisting the coding processing in not only software but also hardware design.
Recently, LLMs have been employed for translating natural language to HDL code~\cite{verilogcoder,verigen}, assisting with debugging~\cite{rtlfixer,hdldebugger}, and optimizing code implementation to improve key chip metrics such as performance, power, and area (PPA) \cite{rtlwriter}. These efforts primarily focus on traditional HDLs like Verilog.

\textit{Chisel}~\cite{chisel-release} 
is a next-generation HDL embedded in Scala that facilitates advanced circuit design. Unlike Verilog, which primarily targets the Register Transfer Level (RTL) and often necessitates detailed low-level specifications, Chisel offers higher-level abstractions through object-oriented and functional programming features. This approach results in more concise, maintainable, and scalable designs~\cite{chisel-whitepaper}. Due to its compatibility with agile design methodologies, Chisel is gaining popularity for complex hardware design~\cite{chisel-compare},
including processors (e.g., XiangShan~\cite{xiangshan}, BOOM~\cite{boom}, and RocketChip~\cite{asanovic2016rocket}), 
accelerators (e.g., MAGMA-Si~\cite{magma-si} and Gemmini~\cite{genc2019gemmini}), among many others.

However, the potential of using LLMs for Chisel code generation remains largely unexplored. As a relatively new language, Chisel has significantly less publicly available code online—for instance, the amount of Chisel code on GitHub is only about 1/20th that of Verilog—which limits the training data available for LLMs to generate Chisel code effectively. To better understand these limitations, we conducted experiments to evaluate the baseline performance of mainstream LLMs in automatically generating Chisel code in a zero-shot setting. The experiment results show that the baseline capabilities of these LLMs are indeed limited.

In this paper, we propose \oursys, an LLM-based agentic system designed to improve the effectiveness of using LLM for Chisel code generation.
ReChisel incorporates a reflection mechanism, utilizing an iterative process to refine the quality of generated code. The key insight is leveraging feedback from compilation and simulation processes to guide the LLM in correcting errors. Additionally, we observed that during the reflection process, the LLM may enter a non-progress loop. To overcome this, we developed an escape mechanism that enables the LLM to break free from these loops and resume making useful attempts. 
We evaluate the performance of \oursys\ on three benchmarks,  VerilogEval's Spec-to-RTL \cite{verilogeval2}, AutoChip’s HDLBits \cite{hdlbits} and RTLLM~\cite{rtllm2}, with five mainstream LLMs (various versions of GPT-4 and Claude 3.5).
The experimental results show that our method significantly improves the success rate of generating Chisel code. 
Additionally, we compare the performance of LLMs in generating Chisel versus Verilog code. Although the baseline capabilities of LLMs are significantly weaker for Chisel code generation than for Verilog, the performance of \oursys\ reaches a level comparable to the state-of-the-art LLM-based agentic systems for Verilog code generation. This not only validates the effectiveness of our proposed method but also highlights the potential of Chisel in LLM-assisted hardware design. \oursys\ is open-sourced at \url{https://github.com/niujuxin/rechisel}.

%% file: z-background.tex
\section{Background \& Related Work}

\subsection{LLM-assisted Hardware Design}

In hardware design, coding with hardware description languages (HDLs) like Verilog is a time-consuming and labor-intensive task. Recent advancements in LLMs have demonstrated transformative potential in diverse domains, sparking increasing interest in leveraging LLMs to streamline and improve the hardware design process.

Some research~\cite{chipgpt} aims at automating the end-to-end hardware design process using LLMs. 
ChatCPU~\cite{chatcpu} introduces the first agile hardware design and verification platform, and develops a 6-stage pipelined RISC-V CPU prototype with the platform.
Other studies~\cite{verigen,chipchat,vgv,dave} focus on enhancing the effectiveness of HDL code generation through LLMs. 
RTLCoder~\cite{rtlcoder} presents an agent-based framework for automatically generating labeled datasets, which are then used to fine-tune a lightweight model that achieves performance comparable to GPT models. 
RTLRewriter~\cite{rtlwriter} leverages LLMs for code optimization to enhance key chip metrics such as power, performance, and area (PPA).
Some work~\cite{thakur2023benchmarking} emphasizes assessing the performance of LLMs in hardware design tasks. 
RTLLM~\cite{rtllm,rtllm2} provides an open-source RTL generation benchmark consisting of 50 design problems that span a diverse and broad spectrum of complexities. 
VerilogEval~\cite{verilogeval,verilogeval2} provides benchmarks for two tasks: code completion and translating specifications to RTL. 
MG-Verilog dataset~\cite{mgverilog} divides module implementations into multiple abstraction levels, enabling the evaluation of LLM capabilities across varying degrees of abstraction.
These works focus on Verilog, leaving the effectiveness of using LLMs for Chisel code generation an open question yet to be explored.

\subsection{Chisel: Next-Generation HDL}

\textit{Chisel}~\cite{chisel-release} is a next-generation HDL embedded in Scala, leveraging Scala’s advanced syntax and features to enable high-level circuit design. Unlike traditional HDLs like Verilog, which focus on register-transfer level (RTL) and require detailed low-level specifies, Chisel introduces modern programming paradigms such as object-oriented design (OOD) and functional programming. 
This allows for modular, reusable, and parameterized hardware design~\cite{chisel-compare}.
 Chisel is gaining popularity for complex hardware design~\cite{chisel-compare},
including processors (e.g., XiangShan~\cite{xiangshan}, BOOM~\cite{boom}, and RocketChip~\cite{asanovic2016rocket}), 
accelerators (e.g., MAGMA-Si~\cite{magma-si} and Gemmini~\cite{genc2019gemmini}), among many others.
ChatChisel~\cite{chatchisel} leverages LLMs to generate Chisel code, designing and implementing a RISC-V CPU with a five-stage pipeline and branch prediction. This work shows the advantages of using LLMs with Chisel in complex hardware design scenarios, highlighting Chisel’s potential of using LLM to assist in hardware generation.

Chisel is not directly used for hardware implementation; instead, it serves as a high-level language that must first be \textit{compiled} into Verilog~\cite{chisel-book}.
The compilation process begins with the execution of Chisel code, using the Chisel library to construct a hardware design graph.
This design is then serialized into a platform-independent intermediate representation known as FIRRTL~\cite{firrtl,firrtl-report}. 
The FIRRTL representation is optimized and ultimately transformed into Verilog code through the FIRRTL compiler.

\subsection{Reflection}

\textit{Reflection}~\cite{reflexion,react,chain-of-thought} enables LLMs to self-analyze their reasoning processes and actions, dynamically adapting to enhance reliability, adaptability, and effectiveness.
In the context of code generation, reflection allows LLMs to automatically identify syntax errors or logical flaws in generated code.
AutoChip~\cite{autochip} leverages reflection to improve the performance of Verilog code generation. 
VerilogCoder~\cite{verilogcoder} introduced an autonomous agent capable of performing graph planning, and integrates an AST-based waveform tracing tool to provide feedback for Verilog code debugging.

%% file: z-motivation.tex
\section{Motivation}
\label{sec:motivation}

\begin{table}
\centering
\caption{Comparison of LLM baseline capabilities in generating Chisel and Verilog code.}
\label{tab:baseline-eval}
\renewcommand{\arraystretch}{1.} 
\begin{tabular}{ccccccc}
\Xhline{1pt}
\multirow{2}{*}{\textbf{Model}} & \multicolumn{2}{c}{\textbf{Pass@1 (\%)}} & \multicolumn{2}{c}{\textbf{Pass@5 (\%)}} & \multicolumn{2}{c}{\textbf{Pass@10 (\%)}} \\
 & \textbf{CHS} & \textbf{VRL} & \textbf{CHS} & \textbf{VRL} & \textbf{CHS} & \textbf{VRL} \\ \hline
GPT-4 Turbo & \cellcolor{red!10}45.54 & \cellcolor{green!10}67.61 & \cellcolor{red!10}61.97 & \cellcolor{green!10}77.46 & \cellcolor{red!10}66.20 & \cellcolor{green!10}81.22 \\
GPT-4o & \cellcolor{red!10}45.07 & \cellcolor{green!10}69.48 & \cellcolor{red!10}65.26 & \cellcolor{green!10}75.59 & \cellcolor{red!10}70.89 & \cellcolor{green!10}77.46 \\
GPT-4o mini & \cellcolor{red!10}11.27 & \cellcolor{green!10}59.15 & \cellcolor{red!10}28.64 & \cellcolor{green!10}69.48 & \cellcolor{red!10}36.62 & \cellcolor{green!10}72.30 \\
Claude 3.5 Sonnet & \cellcolor{red!10}33.33 & \cellcolor{green!10}77.93 & \cellcolor{red!10}52.58 & \cellcolor{green!10}82.16 & \cellcolor{red!10}59.62 & \cellcolor{green!10}84.04 \\
Claude 3.5 Haiku & \cellcolor{red!10}26.29 & \cellcolor{green!10}75.59 & \cellcolor{red!10}54.46 & \cellcolor{green!10}83.57 & \cellcolor{red!10}58.69 & \cellcolor{green!10}84.04 \\
\Xhline{1pt}
\multicolumn{7}{p{0.9\columnwidth}}{\footnotesize * CHS and VRL represent Chisel and Verilog, respectively.}
\end{tabular}
\end{table}
\begin{figure}
    \centering
    \includegraphics[width=0.92\columnwidth]{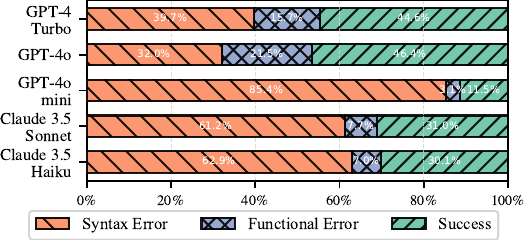}
    \caption{Proportion of different error types in generated Chisel code.}
    \label{fig:baseline-error-type}
\end{figure}

Chisel's high-level abstractions and modern programming principles make it well-suited for agile hardware design and increasingly popular for RTL design. However, it remains unclear whether Chisel is also a viable choice for automatic RTL code generation using LLMs. Should we use LLMs to directly generate Verilog code from a natural language specification, or should we generate Chisel code and then compile it into Verilog? 
As a first step towards answering this question, this paper explores the capability of LLMs to generate \emph{correct} Chisel code.

Compared to Verilog, Chisel presents both potential advantages and disadvantages for LLM-based automatic code generation. 
On the downside, Chisel is relatively new and has significantly less publicly available code for model training—approximately only $1/20^{\texttt{th}}$ the volume of Verilog code on GitHub
\footnote{As of March 2025, our GitHub advanced search revealed 1.60 million lines of publicly available Verilog/SystemVerilog code and 80.1 thousand lines of Scala code importing Chisel extensions.}. 
On the upside, Chisel's similarity to software languages and its incorporation of modern programming concepts enables LLMs to focus on higher-level logic in complex scenarios rather than low-level specifications. Additionally, the strong capabilities of LLMs in handling software code like Java could potentially transfer to Chisel code generation. Given these factors, it remains uncertain whether Chisel can be a competitive choice for automatic RTL code generation using LLMs.

%


As a preliminary step toward understanding LLMs’ performance in Chisel code generation, we assessed mainstream LLMs’ \textit{baseline} capabilities in generating Chisel code, and compared it with the more widely studied Verilog to identify potential performance gaps.
The evaluation was performed under \textit{zero-shot} conditions, providing the LLMs with only the specification without additional context. Each model was tasked with producing \textit{complete} module implementations in either Verilog or Chisel in a \textit{single} pass. 
We evaluate using the widely adopted Pass@$k$~\cite{passk} metric, which measures the probability of achieving at least one correct result among the top $k$ generated attempts.
Results in Table~\ref{tab:baseline-eval} show that the ability of LLMs in HDL code generation to use Chisel is significantly less effective than Verilog.
We further analyze the proportions of different types of errors that occur when LLMs generate Chisel code.
As illustrated in Fig.~\ref{fig:baseline-error-type}, a significant proportion of errors arise during the compilation phase, which indicates that these LLMs still struggle with ensuring the syntactic correctness of Chisel code.
This huge performance gap highlights the necessity of exploring new methods to enhance LLMs' ability to generate Chisel code.


%% file: z-new-design.tex
\section{\oursys~Design}
\label{sec:design}

\begin{figure}
    \centering
    \includegraphics[width=.95\columnwidth]{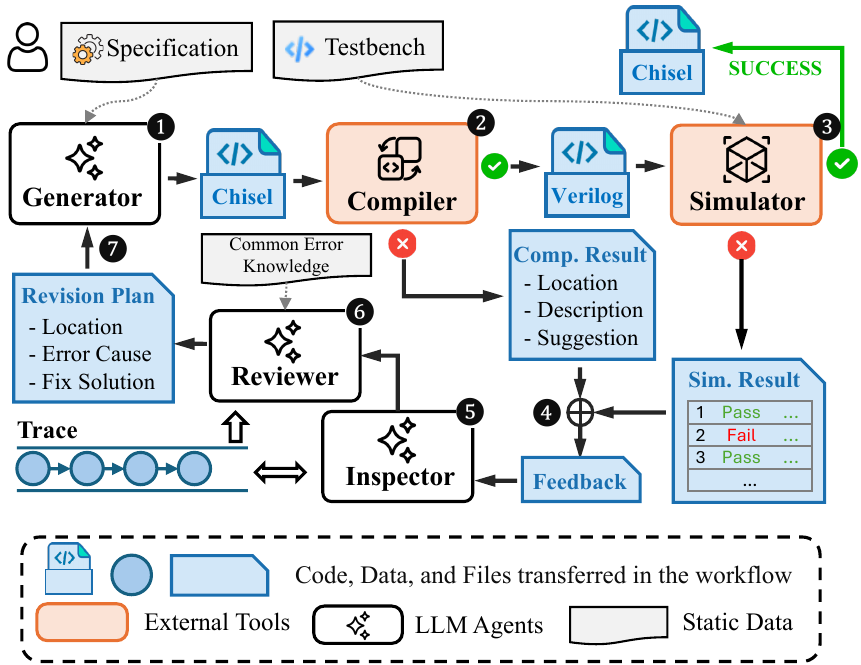}
    \caption{\oursys~workflow: \oursys~is an LLM-based agentic system designed to enhance the effectiveness of Chisel code generation.
    }
    \label{fig:system-design}
\end{figure}




\subsection{Overview}

We introduce \oursys, an LLM-based agentic system designed to enhance the effectiveness of Chisel code generation. 
Fig.~\ref{fig:system-design} illustrates the workflow of \oursys. Given a specification and a testbench, the goal is to generate correct Chisel code that passes all tests. 

The workflow begins with the \textit{Generator} creating Chisel code based on the specification (\stepa). The \textit{Compiler} then translates this code into Verilog (\stepb), followed by functional testing of the Verilog code using the \textit{Simulator} (\stepc). 
If both compilation and simulation succeed, the Chisel code is considered correct, concluding the workflow, and the user receives the validated code.
However, if errors occur, \oursys~employs a reflection mechanism to iteratively generate code until it succeeds. 
In each iteration, results are collected and organized to provide feedback to the \textit{Inspector} (\stepd). The Inspector maintains a trace that includes all previous iterations. Upon receiving the feedback, the Inspector updates the trace (\stepe). The \textit{Reviewer} then analyzes the trace to formulate a revision plan (\stepf), which is passed back to the \textit{Generator} as guidance for error correction. The Generator applies the necessary code adjustments according to the plan to produce a new version of the Chisel code (\stepg). The updated code re-enters the workflow, initiating the next iteration. 
%
The Generator, Reviewer, and Inspector are LLM agents whose roles in the workflow are defined by the provided system prompts, while the Compiler and the Simulator are external tools.

A \textit{maximum number of iterations} is set to limit the number of attempts the LLM can make. 
If the number of iterations exceeds this limit, the workflow terminates, and the case is considered a failure.
The number of iterations is a key factor affecting the effectiveness of \oursys. 
In general, a higher number of iterations typically leads to better performance. 
However, the performance gains from increasing the number of iterations is limited by the inherent capabilities of the LLM. 
In Section~\ref{sec:evaluation}, we provide a detailed analysis of how the number of iterations impacts the effectiveness of \oursys.

To enhance the reflection process, we propose different feedback strategies for different types of results. 
We also adopt in-context learning methods for more effective reviews. 
These approaches are detailed in Section~\ref{sec:design:reflection}.
Additionally, we observed that the LLMs may fall into a \textit{non-progress loop} during reflection. To address this issue, we propose an \textit{escape} mechanism implemented within the Inspector. The detailed discussion is given in Section~\ref{sec:design:escape}.

\subsection{Error Feedback \& Reflection Mechanism}
\label{sec:design:reflection}

Errors in testing LLM-generated Chisel code fall into two types: \textit{syntax errors} reported in compilation and \textit{functional errors} discovered in simulation. 
We propose different feedback strategies for each type.

\begin{figure}
    \centering
    \includegraphics[width=.95\columnwidth]{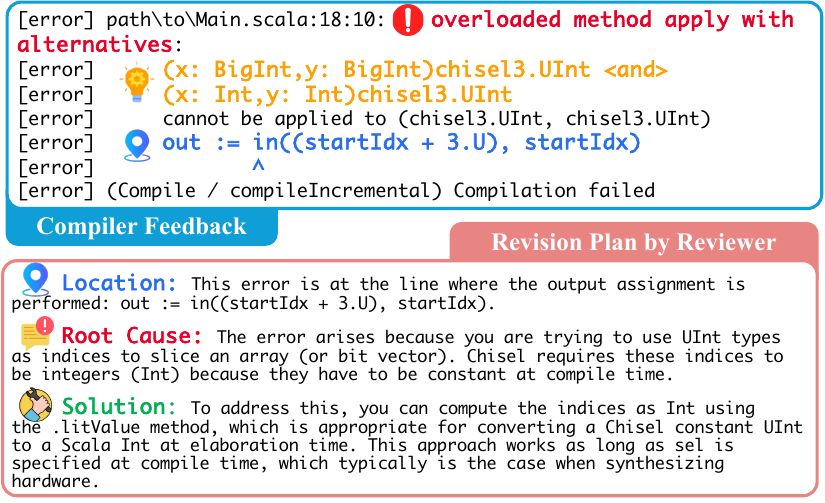}
    \caption{Example of compiler feedback and revision plan.}
    \label{fig:feedback-revision-example}
\end{figure}

\textbf{Syntax Error.}
During the compilation process, the compiler conducts a series of checks to ensure that the code is syntactically correct and can be successfully translated.
When errors are detected, the compiler reports the specific location of each issue, provides a detailed explanation of the error and, in some cases, suggests possible fixes. 
This information is organized into an error list and serves as feedback. Fig.~\ref{fig:feedback-revision-example} illustrates an example of compiler feedback.

\input{chisel-compiler-feedback-table}

We found that the types of syntax errors tend to be densely distributed. 
Specifically, the most common errors generated by the LLM include mixing Scala and Chisel syntax, handling signal types, and managing clock domains. 
In Table~\ref{tab:error-list}, we show several examples of common errors made by the LLM along with the corresponding feedback provided by the compiler.
Beyond syntax errors, we found that the compiler performs static code analysis during compilation and reports logical errors that can cause functional errors in simulations. 
For instance, the compiler can detect uninitialized or partially initialized signals (B3), preventing unintended latch generation in hardware. Besides, it can identify combinational loops (C2) to avoid unpredictable signal oscillations.
For simplicity, we also refer to these logical errors detected at compile time as syntax errors in the future discussion.
For these common errors, we employ in-context learning~\cite{icl-survey} to further enhance the effectiveness of reviews. 
We pre-organized the causes and corresponding fix guidance for each error and included this information within the prompts provided to the LLM. 
This approach enables the LLM to deliver feedback more quickly and accurately when encountering such errors.

\textbf{Functional Error.}
After compiling the Chisel code into Verilog, the Verilog code is passed to the simulator as the Device Under Test (DUT). In addition to the DUT, the simulation process includes a reference module and a series of functional points. Each functional point consists of input stimuli and the expected output.
During simulation, input stimuli are applied to both the DUT and the reference module, and their outputs are compared. If the outputs match, the test passes; otherwise, it fails.
For all failed test points, we extract the input signals, the expected output signals, and the actual obtained signals to create an error list, which is used as feedback.

Feedback for the two aforementioned error types is provided to the LLM alongside the Chisel code for reflection.
The LLM generates a revision plan that, for each error, includes the error location, a cause analysis, and specific solutions. 
Fig.~\ref{fig:feedback-revision-example} also illustrates a revision plan based on the compiler feedback shown in the figure. 
The revision plan guides the LLM in making adjustments to the Chisel code.

\subsection{Non-progress Loop \& Escape Mechanism}
\label{sec:design:escape}

\begin{figure}
    \centering
    \includegraphics[width=.9\columnwidth]{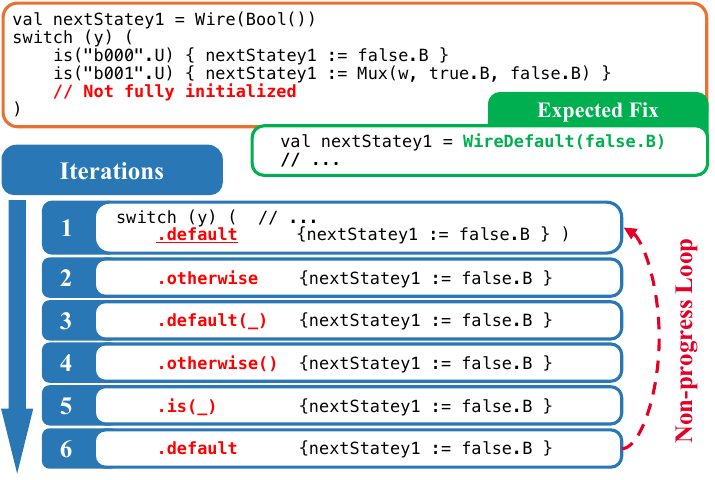}
    \caption{Example of non-progress loop. The LLM is expected to provide default values when defining the signal to ensure fully initialization. However, it repeatedly attempts to add default statements within the \texttt{switch} block.}
    \label{fig:deadlock}
\end{figure}

During the iterative process of code optimization through reflection, LLMs may fall into a \textit{non-progress loop}, i.e., the LLM repeatedly cycles through the same set of errors without resolving them, and is unable to break free from the error cycle.
We use Fig.~\ref{fig:deadlock} to illustrate this problem.
The code error lies within the \texttt{switch} block where not all possible cases of \texttt{y} are handled, leading to \texttt{nextStatey1} being uninitialized when \texttt{y} takes unspecified values. 
The correct approach to fix this is to use \texttt{WireDefault} when defining the signal to provide a default value. 
However, in the first reflection iteration, the LLM tried to add a default case to the \texttt{switch} block incorrectly, as Chisel’s \texttt{switch} does not support any default case syntax. 
Subsequently, in each iteration, the LLM fell into the trap of repeatedly attempting to add a default case to the \texttt{switch} block, resulting in a non-progress loop.

The non-progress loops happen mainly due to the greedy strategy during reflection and the limitations of the LLM’s context.
In the example in Fig.~\ref{fig:deadlock}, after the first iteration fails, the compiler issues an error message indicating unsupported syntax in the \texttt{switch} block. Affected by this error, the reviewer focuses solely on correcting issues inside the \texttt{switch} block in subsequent attempts, ignoring the true location of the error which can only be captured from a global perspective.
Additionally, the LLM has a limited context length, which means that as the number of iterations increases, early error experiences are completely forgotten, leading the LLM to return to its previous state after multiple attempts.

\begin{figure}
    \centering
    \includegraphics[width=0.92\columnwidth]{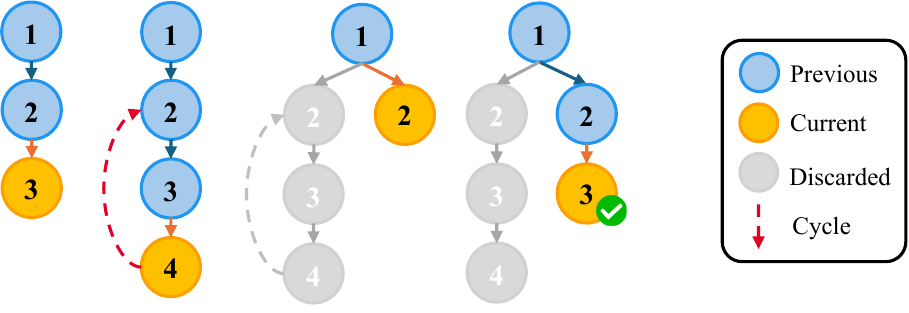}
    \caption{Illustration of escape mechanism.}
    \label{fig:escape}
\end{figure}

To address the non-progress loop issue, we propose an \emph{escape} mechanism that monitors the LLM’s iterative process for potential cycles. 
During each iteration, the trace and the current feedback are provided to the Inspector for analysis to detect any cycles. 
Specifically, the feedback is compared with each entry in the trace. 
If the current feedback contains an error at the same location as a previous entry’s error, and the LLM determines that the causes of these errors are identical, then all iterations between these two points form a non-progress loop. 
In such cases, the LLM discards all iterations involved in the loop. The Reviewer then regenerates a revision plan starting from the step immediately preceding the loop. With the inherent diversity, the LLM is expected to break out of the loop and potentially find the correct fix method. Fig.~\ref{fig:escape} illustrates this escape mechanism.

%% file: chisel-compiler-feedback-table.tex
\begin{table*}
\centering
\caption{Common syntax errors in LLM-generated Chisel code and corresponding compiler feedback.}
\renewcommand{\arraystretch}{1} 
\begin{tabular}{p{0.22\textwidth}p{0.18\textwidth}p{0.20\textwidth}p{0.30\textwidth}}
\Xhline{1pt}
\small\textbf{Description} & 
\makecell[c]{\small\textbf{Incorrect}} & 
\makecell[c]{\small\textbf{Corrected}} & 
\makecell[c]{\small\textbf{Compiler Feedback}} \\ \Xhline{1pt}
\multicolumn{4}{c}{\textbf{A. Structural Errors}} \\ \hline \hline 
\textbf{1. Misspelling, unmatched parentheses, etc.} & 
\cellcolor{red!10}\makecell[lt]{\ttf val signal=Wire(...)\\\ttf \wcode{sgnal}:=0.U} & 
\cellcolor{green!10}\makecell[lt]{\ttf val signal=Wire(...)\\\ttf \tcode{signal}:=0.U} & 
\scriptsize\texttt{\fdbk{Value sgnal is not a member.} \corr{Did you mean signal?}} \\ \hline
\textbf{2. Mixed usage of Chisel and Scala syntax.} & 
\cellcolor{red!10}\makecell[lt]{\ttf out := \\\ttf \wcode{in.asInstanceOf[SInt]}} & 
\cellcolor{green!10}\makecell[lt]{\ttf out := \tcode{in.asSInt}} & 
\scriptsize\texttt{\fdbk{class chisel3.UInt cannot be cast to class chisel3.SInt}.} \\ \hline
\textbf{3. Incorrect invocations of functions or methods.} & 
\cellcolor{red!10}\makecell[lt]{\ttf val r=Seq.fill(5)(...) \\\ttf \wcode{r(0, 2)} } & 
\cellcolor{green!10}\makecell[lt]{\ttf val r=Seq.fill(5)(...) \\\ttf \tcode{r(2)} } & 
\scriptsize\texttt{\fdbk{Too many arguments.} \corr{Found 2, expected 1 for method apply:(i:Int)}}
\\ \hline
\hline \multicolumn{4}{c}{\textbf{B. Signal Definition, Usage and Typing Errors}} \\ \hline \hline
\textbf{1. Incorrect definition of clock or reset signals.} & 
\cellcolor{red!10}\makecell[lt]{\ttf val rst = \\\ttf IO(Input(\wcode{Reset()}))} & 
\cellcolor{green!10}\makecell[lt]{\ttf val rst = \\\ttf IO(Input(\tcode{Bool()}))} & 
\scriptsize\texttt{\fdbk{A port rst with abstract reset type was unable to be inferred by InferResets.}}
\\ \hline
\textbf{2. Failure to encapsulate interface signals within \ttf IO().} & 
\cellcolor{red!10}\makecell[lt]{\ttf val clk=\wcode{Input(Clock())}} & 
\cellcolor{green!10}\makecell[lt]{\ttf val clk=\tcode{IO(Input(Clock))}} & 
\scriptsize\texttt{\fdbk{Clock must be hardware, not a bare Chisel type}. \corr{Perhaps you forgot to wrap it in Wire(\_) or IO(\_)?}}
\\ \hline 
\textbf{3. Wire signal not (fully) initialized.} & 
\cellcolor{red!10}\makecell[lt]{\ttf when (io.in) \{w:=0.U\} \\\ttf \wcode{// ?}} & 
\cellcolor{green!10}\makecell[lt]{\ttf when (io.in) \{w:=0.U\} \\\ttf \tcode{otherwise \{w:=1.U\}} } & 
\scriptsize\texttt{\fdbk{Reference w not fully initialized.}}
\\ \hline
%
%
%
%
\textbf{4. Bundle connection mismatch.} & 
\cellcolor{red!10}\makecell[lt]{\ttf val a = OneBdl(...) \\\ttf val b = AnotherBdl(...) \\\ttf \wcode{a := b} } & 
\cellcolor{green!10}\makecell[lt]{\\\ttf \tcode{// assignment not allowed} } & 
\scriptsize\texttt{\fdbk{Connection between sink (... OneBdl) and source (... AnotherBdl) failed}: \corr{.cSource Record missing field (c)}.}
 \\ \hline
\textbf{5. Signal type mismatch.} & 
\cellcolor{red!10}\makecell[lt]{\ttf val cnt=oks.\\\ttf reduce(\wcode{\_ +\& \_})} & 
\cellcolor{green!10}\makecell[lt]{\ttf val cnt=oks.\tcode{map(\_.asUInt)}.\\\ttf reduce(\tcode{\_ +\& \_})} &
\makecell[lt]{ \scriptsize\texttt{\fdbk{found: chisel3.Bool}} \\ \scriptsize\texttt{\corr{required: chisel3.UInt}}}
 \\ \hline
\textbf{6. Unsupported signal type conversion or casting.} & 
\cellcolor{red!10}\makecell[lt]{\ttf invertedClk := \\\ttf ($\sim$ clk.asUInt)\wcode{.asClock}} & 
\cellcolor{green!10}\makecell[lt]{\ttf \tcode{// unsupported}} & 
\scriptsize\texttt{\fdbk{Value asClock is not a member of chisel3.UInt.}}
 \\ \hline
\textbf{7. Out-of-bounds access on array-type signal.} & 
\cellcolor{red!10}\makecell[lt]{\ttf vector\wcode{(-1)} := 0.U } & 
\cellcolor{green!10}\makecell[lt]{\ttf vector\tcode{(0)} := 0.U } & 
\scriptsize\texttt{\fdbk{-1 is out of bounds.} (\corr{min 0, max 3})}
 \\ \hline
\hline \multicolumn{4}{c}{\textbf{C. Miscellaneous Errors}} \\ \hline \hline
\makecell[lt]{\textbf{1. Incorrect Clock Domain} \\
* (in multi-clock design)}
  & 
\cellcolor{red!10}\makecell[lt]{\ttf val out=\wcode{RegNext(in)}} & 
\cellcolor{green!10}\makecell[lt]{\ttf val out=\tcode{withClock(clk)} \\\ttf \tcode{\{} RegNext(in)\tcode{\}} } & 
\scriptsize\texttt{\fdbk{No implicit clock.}}
\\ \hline 
\textbf{2. Combinational Loop} & 
\cellcolor{red!10}\makecell[lt]{\ttf val a=Wire(...) \\\ttf \wcode{a := a + 1.U}} & 
\cellcolor{green!10}\makecell[lt]{\\\ttf \tcode{// remove loop}} & 
\scriptsize\texttt{\fdbk{Detected combinational cycle in a FIRRTL module.} \corr{Sample path: $\{\text{a} \leftarrow \text{a\_T\_1} \leftarrow \dots \leftarrow \text{a\_T} \leftarrow \dots \leftarrow \text{a}\}$.}}
\\ \hline
%
\Xhline{1pt}
\end{tabular}
\label{tab:error-list}
\end{table*}

%% file: z-evaluation.tex
\section{Evaluation}\label{sec:evaluation}

This section is organized as follows: Section~\ref{sec:evaluation:setup} provides the experimental setup. Section~\ref{sec:evaluation:perf} evaluates the performance of \oursys. Section~\ref{sec:evaluation:compare} compares \oursys\ with AutoChip, an LLM-based automatic Verilog code generation system. Finally, Section~\ref{sec:evaluation:case} presents a case study to demonstrate the workflow of \oursys.

\subsection{Experimental Setup}
\label{sec:evaluation:setup}

\textbf{Benchmark.}
The evaluation is conducted at the module level, where each test case comprises a module specification that includes a functional description and definitions of I/O signals. 
Our benchmark specifications are sourced from the VerilogEval's Spec-to-RTL \cite{verilogeval2}, AutoChip’s HDLBits \cite{hdlbits}, and RTLLM~\cite{rtllm2} dataset. While most test cases align with our evaluation criteria, some are invalid or incompatible. In particular, issues arise when: 
    1) the case requires using certain Verilog features such as the parameterization\footnote{Chisel employs parameterized module generators as a replacement for the \texttt{\#(parameter ...)} mechanism in Verilog.}, which are not supported by Chisel. As a result, it is impossible to write Chisel code that compiles into Verilog code incorporating these features; 
    2) the correctness of generated code cannot be verified due to missing or incorrect reference code; or 
    3) the case is designed specifically for Verilog code debugging or completion and cannot be adapted for Chisel code generation tasks.
After filtering out these cases, we finalized a set of 216 valid test cases.

\textbf{Model.}
Five mainstream LLMs are selected for evaluation: 
GPT-4 Turbo (version: 2024-04-09), 
GPT-4o (version: 2024-08-06), 
GPT-4o mini (version: 2024-07-18), and
Claude 3.5 Sonnet (version: 2024-10-22), 
Claude 3.5 Haiku (version: 2024-10-22).

\textbf{Evaluation Parameters.}
LLMs are probabilistic, meaning the same prompt can generate different outputs in multiple attempts. 
Therefore, each case is tested ten times, and the Pass@$k$~\cite{passk} metric is calculated. 
The LLM is evaluated in its default configuration without manual adjustments to parameters such as temperature and top-$p$.
The maximum number of iterations is capped at 10.

\subsection{Evaluation on \oursys}
\label{sec:evaluation:perf}

\input{eval-overview}

Table~\ref{tab:baseline-eval} demonstrates the performance improvements achieved by \oursys. 
The results indicate that \oursys~significantly enhances the performance of LLMs in generating Chisel code, with improvements ranging from $10\%$ to $50\%$. The four LLMs—GPT-4 Turbo, GPT-4o, Claude 3.5 Sonnet, and Claude 3.5 Haiku—exhibit very similar peak performances, while GPT-4o mini’s performance is notably inferior. Among these, Claude 3.5 Sonnet achieved the highest final success rates, with $84.98\%$ under Pass@$1$ and $93.43\%$ under Pass@$10$. 
Notably, although Claude 3.5 Haiku is much more lightweight than Sonnet, its performance is nearly on par with that of Sonnet, and it even surpasses GPT-4o and GPT-4 Turbo.

\begin{figure*}
    \centering
    \includegraphics[width=.98\linewidth]{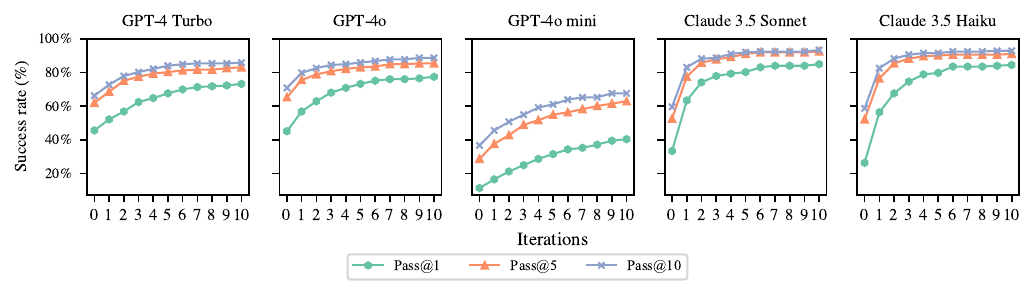}
    \caption{Relationship between success rate and number of iterations for different LLMs.}
    \label{fig:pass-at-k-iteration}
\end{figure*}

Fig.~\ref{fig:pass-at-k-iteration} further illustrates how the success rates of LLMs change as the number of iterations increases.
As shown in the figure, the two Claude LLMs demonstrate much stronger reflection capabilities. 
Taking Sonnet as an example, under the Pass@$1$ metric, its baseline accuracy ($n=0$) is only $33\%$, significantly lower than that of GPT-4 Turbo or GPT-4o. However, after a few iterations, its performance improves substantially, eventually surpassing GPT-4~Turbo and GPT-4o at $n=10$. 
After about four iterations, all LLMs except GPT-4o mini can reach a performance plateau. In contrast, GPT-4o~mini’s performance improves more slowly as the number of iterations increases. 
After reaching the plateau, about $10\%$ of cases still cannot be generated correctly, highlighting the inherent limitations of LLMs.

\begin{figure}
    \centering
    \includegraphics[width=0.92\columnwidth]{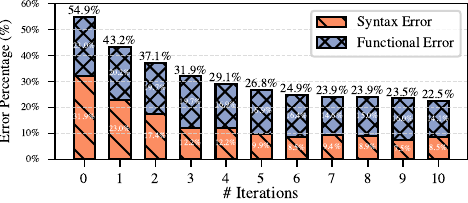}
    \caption{Proportion of syntax and functional errors across iterations. Example of GPT-4o under Pass@$1$ metric.}
    \label{fig:type-error-through-iter}
\end{figure}

We also examine how the number of syntax and functional errors changes with increasing iterations. 
In Fig.~\ref{fig:type-error-through-iter}, we present the proportions of syntax and functional errors for GPT-4o under the Pass@$1$ metric across each iteration. 
Compared to the baseline, \oursys\ reduces both syntax and functional errors. However, we observed that in some cases (e.g., from the 9\textsuperscript{th} to the 10\textsuperscript{th} iteration in the figure), the proportion of syntax errors increased. This suggests that when the LLM fixes functional errors, it may inadvertently reintroduce syntax errors.

\subsection{Performance Comparison of \oursys\ and AutoChip}
\label{sec:evaluation:compare}

\input{chisel-vs-verilog}

To further investigate the questions raised in Section~\ref{sec:motivation}, we compare ReChisel with AutoChip~\cite{autochip}. 
AutoChip is an LLM-based agentic system designed for automatic \textit{direct} Verilog code generation. 
Both systems are evaluated using the same benchmarks and parameters given in Section~\ref{sec:evaluation:setup}. 
We select OpenAI’s GPT-4o, GPT-4 Turbo, and Claude 3.5’s Sonnet models for comparison. 
For both systems, the maximum number of iterations is set to 10.
The results presented in Table~\ref{tab:rechisel-vs-autochip} demonstrate that \oursys\ enhances the LLM’s Chisel code generation performance to a level almost {comparable} with Verilog generation. Specifically, for the GPT-4o model, ReChisel {outperforms} AutoChip in both Pass@$5$ and Pass@$10$ metrics.
These results demonstrate the great potential of using Chisel for hardware code generation.

%% file: eval-overview.tex
\begin{table}
\centering
\caption{\oursys~performance}
\label{tab:rechisel-vs-autochip}
\renewcommand{\arraystretch}{1.}
\begin{tabular}{cc|cccc}
\Xhline{1.2pt}
\multirow{3}{*}{\textbf{Metric}} & \multirow{3}{*}{\textbf{LLMs}} & \multicolumn{4}{c}{\textbf{Success Rate (\%)}} \\ 
& & \multicolumn{4}{c}{$\bm{n=}$} \\
& & $\bm0$ & $\bm1$ & $\bm5$ & $\bm{10}$ \\ \hline \hline
\multirow{5}{*}{\textbf{Pass@1}} & GPT-4 Turbo & \cellcolor{green!30}\textbf{45.54} & 52.11 & 67.61 & 73.24 \\
& GPT-4o & \cellcolor{green!10}\cellcolor{green!10}\textbf{45.07} &  \cellcolor{green!10}\textbf{56.81} & 73.24 & 77.46 \\
& GPT-4o mini & 11.27 & 16.43 & 31.46 & 40.38 \\
& Claude 3.5 Sonnet & 33.33 & \cellcolor{green!30}\textbf{63.38} & \cellcolor{green!30}\textbf{80.28} & \cellcolor{green!30}\textbf{84.98} \\
& Claude 3.5 Haiku & 26.29 & 56.34 & \cellcolor{green!10}\textbf{79.81} & \cellcolor{green!10}\textbf{84.51} \\ \hline
\multirow{5}{*}{\textbf{Pass@5}} & GPT-4 Turbo & \cellcolor{green!10}\textbf{61.97} & 68.54 & 80.28 & 83.10 \\
& GPT-4o & \cellcolor{green!30}\textbf{65.26} &  75.59 & 83.10 & 85.45\\
& GPT-4o mini & 28.64 & 37.56 & 54.93 & 62.91 \\
& Claude 3.5 Sonnet & 52.58 & \cellcolor{green!30}\textbf{77.46} & \cellcolor{green!30}\textbf{91.08} & \cellcolor{green!30}\textbf{92.49} \\
& Claude 3.5 Haiku & 52.11 & \cellcolor{green!10}\textbf{76.53} & \cellcolor{green!10}\textbf{90.14} & \cellcolor{green!10}\textbf{91.08} \\  \hline
\multirow{5}{*}{\textbf{Pass@10}} & GPT-4 Turbo & \cellcolor{green!10}\textbf{66.20} & 72.77 & 84.04 & 85.92 \\
& GPT-4o & \cellcolor{green!30}\textbf{70.89} & 79.81 & 85.92 & 88.73 \\
& GPT-4o mini & 36.62 & 45.54 & 61.03 & 67.61 \\
& Claude 3.5 Sonnet & 59.62 & \cellcolor{green!30}\textbf{83.10} & \cellcolor{green!30}\textbf{92.02} & \cellcolor{green!30}\textbf{93.43} \\ 
& Claude 3.5 Haiku & 58.69 & \cellcolor{green!10}\textbf{82.63} & \cellcolor{green!10}\textbf{91.55} & \cellcolor{green!10}\textbf{92.96} \\
\Xhline{1.2pt}
\multicolumn{6}{l}{* $n$ represents the maximum allowed number of iterations.} \\
\multicolumn{6}{p{0.95\columnwidth}}{** The top-1 and top-2 entries in each metric and $n$ group is highlighted using dark and light green shading, respectively.}
\end{tabular}
\end{table}

%% file: chisel-vs-verilog.tex
\begin{table}
\centering
\caption{Performance comparison of \oursys\ and AutoChip.}
\renewcommand{\arraystretch}{1.}
\begin{tabular}{cc|cccc}
\Xhline{1.2pt}
\multirow{2}{*}{\textbf{Metric}} & \multirow{2}{*}{\textbf{LLMs}} & \multicolumn{2}{c}{\textbf{Success Rate (\%)}} \\ 
& & \oursys & AutoChip \\ \hline \hline
\multirow{3}{*}{\textbf{Pass@1}} & GPT-4 Turbo & \cellcolor{red!20}73.24 & \cellcolor{green!20}\textbf{79.81} \\
& GPT-4o & \cellcolor{red!20}77.46 &  \cellcolor{green!20}\textbf{78.40}  \\
& Claude 3.5 Sonnet & \cellcolor{red!20}84.98 & \cellcolor{green!20}\textbf{91.08}  \\
\hline
\multirow{3}{*}{\textbf{Pass@5}} & GPT-4 Turbo & \cellcolor{red!20}83.10 & \cellcolor{green!20}\textbf{87.79}  \\
& GPT-4o & \cellcolor{green!20}\textbf{85.45} &  \cellcolor{red!20}84.51 \\
& Claude 3.5 Sonnet & \cellcolor{red!20}92.49 & \cellcolor{green!20}\textbf{96.71} \\
\hline
\multirow{3}{*}{\textbf{Pass@10}} & GPT-4 Turbo & \cellcolor{red!20}85.92 & \cellcolor{green!20}\textbf{89.20} \\
& GPT-4o & \cellcolor{green!20}\textbf{88.73} & \cellcolor{red!20}87.79  \\
& Claude 3.5 Sonnet & \cellcolor{red!20}93.43 & \cellcolor{green!20}\textbf{97.65} \\ 
\Xhline{1.2pt}
\end{tabular}
\end{table}

%% file: z-case-study.tex
\subsection{Case Study}
\label{sec:evaluation:case}

\begin{figure}
    \centering
    \includegraphics[width=\columnwidth]{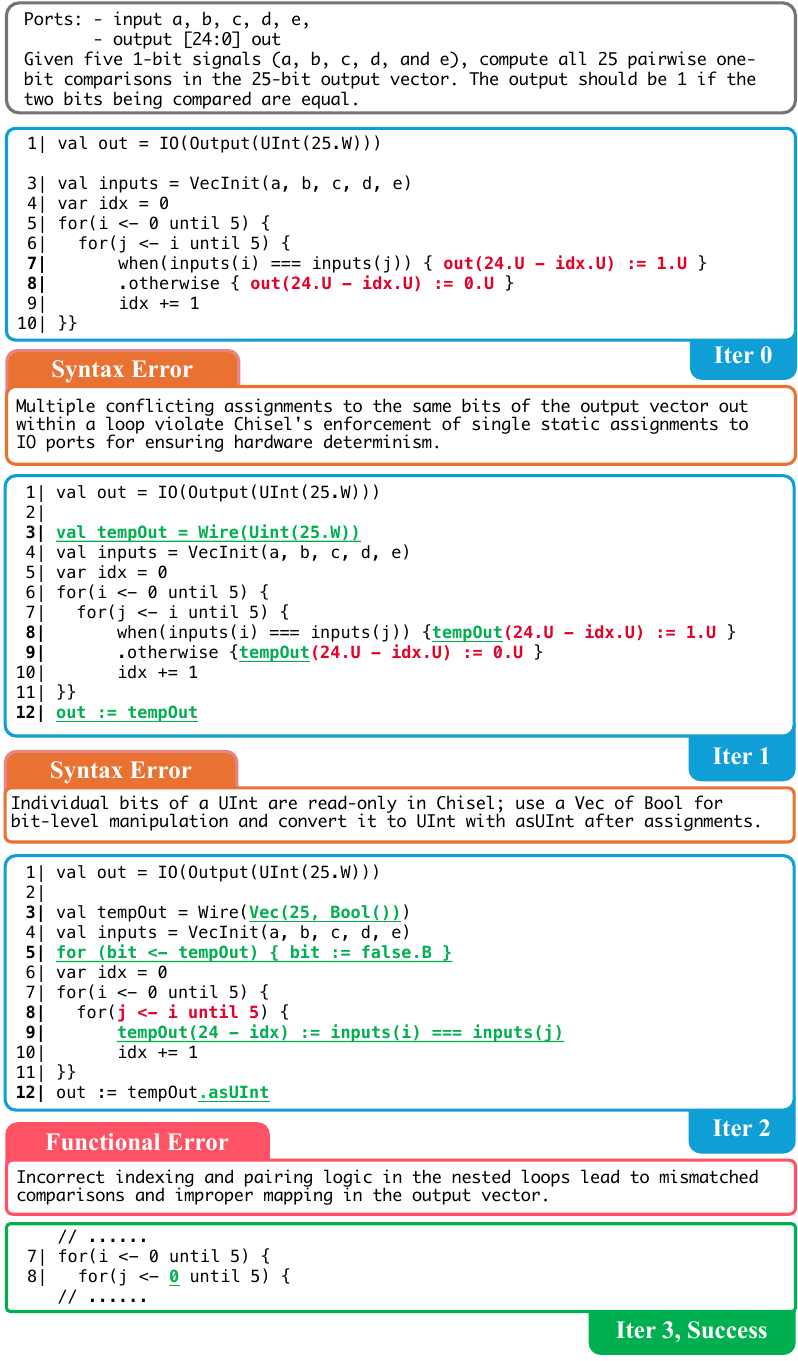}
    \caption{Case study on reflection. Green underlined codes represent modifications based on feedback from the previous iteration, while red codes indicate syntax or functional errors in the current iteration. The content within the error boxes summarizes the corresponding reviewer plans.}
    \label{fig:case-study}
\end{figure}

Fig.~\ref{fig:case-study} illustrates the \oursys's reflection workflow through a case study. 
This case corresponds to the \verb|Vector5| test case from AutoChip’s HDLBits benchmark, utilizing the GPT-4o model.
The figure illustrates the specifications and demonstrates how an LLM iteratively refines its output to progressively generate syntactically correct and functionally accurate Chisel code through reflection. 
The process was completed in three iterations. 
During the first two iterations, the code encountered syntax errors. 
In the first iteration, the LLM introduced intermediate variables to address multiple assignment issues, but an incorrect data type choice caused another error. 
In the second iteration, the LLM used boolean vectors that support bitwise operations, successfully generating syntactically correct code. 
In the third iteration, the syntactically correct code contained a functional error. The LLM then corrected the loop logic, ultimately producing a functionally correct implementation that passed all tests.